\documentclass{article}
\pdfoutput=1

\usepackage[numbers]{natbib}

\usepackage[preprint]{neurips_2022}

\usepackage[utf8]{inputenc} 
\usepackage[T1]{fontenc}    
\usepackage{hyperref}       
\usepackage{url}            
\usepackage{booktabs}       
\usepackage{amsfonts}       
\usepackage{nicefrac}       
\usepackage{microtype}      
\usepackage{xcolor}         

\usepackage{amsmath}
\usepackage{amssymb}
\usepackage{mathtools}
\usepackage{amsthm}

\usepackage{hyperref}
\usepackage{url}
\usepackage{graphicx}
\usepackage{epstopdf}
\usepackage{bbm}
\usepackage{bm}
\usepackage{dsfont}
\usepackage{amssymb}
\usepackage{multirow}
\usepackage{subcaption}

\theoremstyle{plain}
\newtheorem{theorem}{Theorem}[section]
\newtheorem{proposition}[theorem]{Proposition}

\theoremstyle{definition}
\newtheorem{definition}[theorem]{Definition}

\theoremstyle{remark}

\title{Robust Long-Tailed Learning via \\ Label-Aware Bounded CVaR}

\author{%
  Hong Zhu\thanks{Both authors contributed equally to this research.} \\
  Huawei\\
  \texttt{zhuhong8@huawei.com} \\
  \And
  Runpeng Yu\textsuperscript{$\ast$} \\
  National University of Singapore \\
  \texttt{r.yu@u.nus.edu} \\
  \And
  Xing Tang \\
  Tencent \\
  \texttt{shawntang@tencent.com} \\
  \AND
  Yifei Wang \\
  Peking University \\
  \texttt{yifei\_wang@pku.edu.cn} \\
  \And
  Yuan Fang \\
  Huawei \\
  \texttt{frank.fy@huawei.com} \\
  \And
  Yisen Wang \\
  Peking University \\
  \texttt{yisen.wang@pku.edu.cn} \\
}

\begin{document}

\maketitle

\begin{abstract} \label{abstract}

Data in the real-world classification problems are always imbalanced or long-tailed, wherein the majority classes have the most of the samples that dominate the model training. In such setting, the naive model tends to have poor performance on the minority classes. Previously, a variety of loss modifications have been proposed to address the long-tailed leaning problem, while these methods either treat the samples in the same class indiscriminatingly or lack a theoretical guarantee. In this paper, we propose two novel approaches based on CVaR (Conditional Value at Risk) to improve the performance of long-tailed learning with a solid theoretical ground. Specifically, we firstly introduce a \emph{Label-Aware Bounded CVaR} (LAB-CVaR) loss to overcome the pessimistic result of the original CVaR, and further design the optimal weight bounds for LAB-CVaR theoretically. Based on LAB-CVaR, we additionally propose a \emph{LAB-CVaR with logit adjustment} (LAB-CVaR-logit) loss to stabilize the optimization process, where we also offer the theoretical support. Extensive experiments on real-world datasets with long-tailed label distributions verify the superiority of our proposed methods. 
\end{abstract}


\section{Introduction} \label{introduction}

In real-world classification problems, there always have \emph{long-tailed} label distributions, wherein the minority classes have much fewer samples than the majority classes \cite{van2017devil,liu2019large,cao2019learning,zhang2021deep}. In such long-tailed setting, the majority classes dominate the model training, which results in poor performance on the minority classes \cite{menon2020long,kang2019decoupling}. This is particularly undesirable when the minority classes are important, such as in the applications of cancer diagnosis and credit card fraud detection \cite{xu2020class}. Therefore, how to address the long-tailed learning problem is critical.

The extant methods for long-tailed learning could be divided into 4 categories: 
1) under/over sampling of the inputs \cite{kubat1997addressing,wallace2011class,chawla2009data}, 2) post-hoc corrections of the outputs \cite{fawcett1996combining,collell2016reviving,kim2020adjusting}, 3) training procedures manipulation \cite{qi2020attentional,tang2020long,wang2017learning}, and 4) loss function modifications \cite{cao2019learning,menon2020long,effect-number,xu2020class}. 
Besides using one of them alone, methods are further combined together to complement each other, e.g. the under/over sampling method may be easily combined with the methods in the other three categories \cite{menon2020long}, and the loss modification method may be promoted by the combination with a proper training procedure \cite{qi2020attentional,cao2019learning}. In this paper, we focus on the loss modification methods, which assign weights \cite{huang2016learning,effect-number} or relative margins \cite{cao2019learning,menon2020long,samuel2021distributional} to the training loss aiming to balance the label distribution. For the class-level loss modification methods \cite{huang2016learning,effect-number,xu2020class}, their notable disadvantage is that they do not discriminatingly treat the samples in the same class.  On the other hand, the sample-level loss modification methods \cite{lin2017focal,li2019gradient} usually lack a theoretical guarantee.

In this paper, we propose novel methods for long-tailed learning based on CVaR (Conditional Value at Risk) which could consider the difference between the samples in the same class. As one of the most popular distributionally robust optimization (DRO) methods, CVaR aims to maximize the worst-case performance, and then can naturally improve the model performance on minority classes \cite{boostedCVaR,qi2020attentional}. However, in the classification task in which the model is evaluated by the zero-one loss at test time, the original CVaR and DRO are proved to result in the same model performance with ERM (empirical risk minimization) \cite{hu2018does,boostedCVaR}. Different from the solutions proposed in \cite{hu2018does} and \cite{boostedCVaR}, we propose another novel approach to overcome the pessimistic result of the original CVaR in the classification task. This approach is named as \emph{label-aware bounded CVaR} (LAB-CVaR) which means that the weight bounds are varied with labels. Furthermore, we theoretically give the optimal design of the upper and lower weight bounds for each class aiming to minimize the LAB-CVaR loss. On the other hand, LAB-CVaR is a special re-weighting method. But, the re-weighting strategies themselves may cause difficulties and instability in optimization \cite{cao2019learning,effect-number}. Thus, we additionally propose an approach named  \emph{LAB-CVaR with logit adjustment} (LAB-CVaR-logit) to alleviate the influence of the loss weight in optimization, and theoretically study the mechanism that how the logit adjustment technique can improve the stability in optimization. In summary, our contributions are: 
\begin{itemize}
    \item We propose the LAB-CVaR loss which can overcome the pessimistic result of the original CVaR, and theoretically give its optimal design of the upper and lower weight bounds by aiming to minimize the LAB-CVaR loss.
    \item We propose the LAB-CVaR-logit loss which can alleviate the difficulties and instability in optimization caused by the loss re-weighting, and study its mechanism of how to reduce the influence of the loss weight.
    \item Extensive experiments on publicly available real-world datasets show that our approaches outperform the state-of-the-art algorithms significantly.
\end{itemize}

\section{Related work}\label{related}

As aforementioned, we divide the extant long-tailed learning methods into four categories, and mainly focus on the loss modification category in this paper. Thus, we briefly introduce some representative loss modification methods here.

\textbf{Re-weighting.} Re-weighting methods try to balance the class distribution via assigning weights to the training loss. The vanilla re-weighting method is to weight classes proportionally to the inverse sample size of each class \cite{huang2016learning}. Instead of the inverse sample size, \cite{effect-number} defines the effective number as the expected volume of samples, and proposes to use the inverse effective number to do re-weighting. \cite{xu2020class} proves that there exist performance trade-offs in class-weighted methods, and proposes robust approaches based on CVaR. All the class-level re-weighting methods  do not discriminatingly treat the samples in the same class. Besides those class-level re-weighting methods, there also exists some sample-level re-weighting methods, such as Focal loss \cite{lin2017focal} which down-weights the well-classified samples, and GHM \cite{li2019gradient} which uses gradients to check whether the sample is well-classified. But the sample-level re-weighting methods for long-tailed learning usually lack theoretical foundation \cite{qi2020attentional}. In contrast, our proposed LAB-CVaR is a special re-weighting method in which the loss weight is variable during the adversarial training, and not only considers the differences between the samples in the same class, but also has the theoretical guarantee in the design of weight bounds.  

\textbf{Logit adjustment.} \cite{li2002logit1,masnadi2010logit2} propose to class-wisely shift the margin in the hinge loss. \cite{cao2019learning} points out that re-weighting methods may cause difficulties and instability in optimization, and proposes the label-distribution-aware margin loss in the softmax cross-entropy. \cite{menon2020long} proposes a unified framework for re-weighting methods and post-hoc methods with the help of logit adjustment. Different from them, our LAB-CVaR-logit extends the logit adjustment technique to make the loss consistent with the re-weighted loss, and studies the mechanism why the logit adjustment can improve the re-weighting methods in optimization.

\section{Preliminaries}

\subsection{Classification with long-tailed learning}
In a multi-class classification problem, the input space is denoted as $\mathcal{X} \in \mathbb{R}^d$, and the label space is denoted as $\mathcal{Y}=[L] \doteq \{1,\ldots,L \}$. 
The $n$ labeled training data $\{(x_i,y_i)\}_{i=1}^n$ are sampled from $\mathcal{X} \times \mathcal{Y}$ with the joint distribution $P_{x,y}$. 
A prediction model $f:\mathbb{R}^d \rightarrow \mathbb{R}^L$ usually are trained over $P_{x,y}$ by minimizing the \emph{empirical risk} which is defined as
\begin{equation} \label{ermloss}
    R^\ell(f) = \frac{1}{n}\sum_{i=1}^n \ell(f(x_i),y_i),
\end{equation}
where $\ell(\cdot)$ is a loss function, such as the Cross-Entropy loss and Hinge loss. For classification tasks, we usually care about the misclassification error of the model, which is equal to optimize the zero-one loss: 
\begin{equation}\label{01loss}
    R^{\ell_{0/1}}(f)=\frac{1}{n}\sum_{i=1}^n \ell_{0/1}(f(x_i),y_i)=\frac{1}{n}\sum_{i=1}^n \mathds{1}\{y_i \neq \max_{y'\in [L]} f_{y'}(x_i)\},
\end{equation}
where $\mathds{1}$ is the indicator function.
In Eq. \eqref{01loss}, the misclassification error is calculated over the overall data distribution. However, when the label distribution $P_y$ is highly skewed, there exists the \emph{long-tailed learning} or \emph{class imbalance} problem. Then, the overall error in Eq. \eqref{01loss} is not a suitable measure of model performance in such case, because a trivial model which classifies every sample to the majority label will attain a low misclassification error \citep{menon2020long}. To cope with this, we should separately consider the class-wise  misclassification error to improve the model performance over the minority class. The class-wise misclassification error for the class $j$ is defined as
\begin{equation}
    R_j^{\ell_{0/1}}(f) = \frac{1}{n_j}\sum_{y_i=j} \ell_{0/1}(f(x_i),y_i),
\end{equation}
where $n_j$ is the sample size of the class $j$, and it is obvious that $n=\sum_{j=1}^L n_j$, $R^{\ell_{0/1}}=\sum_{j=1}^L \frac{n_j}{n}R_j^{\ell_{0/1}}$. Without loss of generality, we assume $n_1\leq n_2\leq\cdots\leq n_L$. Further more, the balanced error rate (BER) , which averages each of the per-class error rates, namely $\frac{1}{L}\sum_{j=1}^L R_j^{\ell_{0/1}} (f)$, is widely used to evaluate the model performance \citep{chan1998learning,cao2019learning,menon2020long}.

\subsection{DRO and \texorpdfstring{$\alpha$}-CVaR}
Eq. \eqref{ermloss} is also known as the empirical risk minimization (ERM) loss which implicitly assumes that the training and test data have the same distribution. This assumption usually does not hold in real-world scenarios in which there exist unknown distribution shifts between the training and test data, especially in the long-tailed learning problem, the label distribution is highly skewed in the training data, but is balanced even inversely skewed in the test data. 
The distributionally robust optimization (DRO) is a useful technique to solve the unknown distribution shift problem \cite{rahimian2019distributionally} whose effectiveness has been verified in long-tailed learning \cite{qi2020simple,qi2020attentional}. DRO minimizes the worst-case distribution $Q$ within a ball w.r.t. divergence $D$ around the training distribution $P$. The DRO loss is defined as \cite{boostedCVaR} 
\begin{equation}
    \text{DRO}^\ell(f) = \sup_{Q \ll P}\{ \mathds{E}_Q[ \ell(f(x),y)], D(Q||P)\leq \rho\},
\end{equation}
where $Q \ll P$ means $Q$ is absolute continuous to $P$, i.e. $P(x,y)=0 \Rightarrow Q(x,y)=0$, and $\rho$ is the constraint radius of the distribution ball, which controls the degree of distribution shift. 

Different divergence functions (e.g. Kullback-Leibler divergence, Wasserstein metric) derive different DRO risks. In this work, we focus on one of the most popular DRO risks, the conditional value-at-risk at level $\alpha$ ($\alpha$-CVaR), which is derived from the R\'enyi divergence by $D_\epsilon(P||Q)=\frac{1}{\epsilon-1}\log\int(\frac{dP}{dQ})^\epsilon dQ$ where $\epsilon\rightarrow\infty$ and $\rho=-\log\alpha$ \cite{boostedCVaR}. The $\alpha$-CVaR loss is defined as 
\begin{equation}
    \alpha\text{-CVaR}^\ell(f) = \max_{\bm{w}\in \Delta_n,0\leq w_i \leq \frac{1}{\alpha n}} \sum_{i=1}^n w_i\ell(f(x_i),y_i),
\end{equation}
where $\Delta_n=\{(w_1,\cdots,w_n): \sum_{i=1}^n w_i=1\}\in \mathbb{R}^n$ and $\alpha\in(0,1)$. Because of the constraints of $\sum_{i=1}^n w_i=1$ and $0\leq w_i \leq \frac{1}{\alpha n}$, $\alpha$-CVaR is always trained on the worst $\alpha$ fraction of the training data.

\section{The Proposed Approach}  

\subsection{Label-aware bounded CVaR}
In the long-tailed classification scenario, the minority classes tend to have worse classification accuracy than the majority classes \cite{cao2019learning,menon2020long}. Then DRO and $\alpha$-CVaR which pay more attention to the worst samples are naturally expected to be helpful to improve the model performance over the minority classes. However, \cite{hu2018does} theoretically points out a surprising phenomenon that DRO has the same minimizer as ERM in the classification scenario in which the model is evaluated by zero-one loss at test time. \cite{hu2018does} thinks the reason for this phenomenon is that the range of the test distributions to which the DRO tries to be robust is
too wide, then the group DRO \cite{hu2018does,oren2019distributionally,sagawa2019distributionally} is proposed, which reduces the freedom degree of the distribution shift by forcing the samples in the same group to have the same loss weight. The drawback of group DRO is that it cannot discriminatingly treat the samples in the same group. Meanwhile, \cite{boostedCVaR} also finds that $\alpha$-CVaR has the same pessimistic phenomenon, and thinks the reason is that the classifiers are
deterministic, then proposes the Boosted CVaR Classification framework to learn randomized classifiers to circumvent the problem. The drawback of Boosted CVaR Classification is the low efficiency in its boosting procedure. On the other hand, both the group DRO and Boosted CVaR Classification do not specially design their algorithms for long-tailed learning. 

In this paper, we find that when the bound of the loss weight $w_i$ in CVaR is varied across the labels, rather than all the $w_i$ share the same bound $[0,\frac{1}{\alpha n}]$, the minimizers of CVaR and ERM are no longer the same. So beyond the group DRO and Boosted CVaR Classification, we propose a novel approach to overcome the pessimistic result of $\alpha$-CVaR in the classification task, and the approach is named as \emph{label-aware bounded CVaR} (LAB-CVaR) which means that the weight bounds are label-wisely designed. All the detailed proofs of the theory results in this section are shown in Appendix A.

\begin{definition} \label{lab}
\it{Define the loss of LAB-CVaR as:
\begin{equation} \label{eq-lab}
    \text{LAB-CVaR}^\ell(f) = \max_{\bm{w}\in \Delta_n,\frac{1}{\beta_j n}\leq w_i \leq \frac{1}{\alpha_j n}} \sum_{i=1}^n w_i\ell(f(x_i),y_i),
\end{equation}
where $\alpha_j\in\mathbb{R}^+$ and $\beta_j\in\mathbb{R}^+$ are label-wise, and $y_i=j$. }
\end{definition}

\begin{proposition} \label{inverse}
For zero-one loss, the relationship holds: $\text{LAB-CVaR}^{\ell_{0/1}}(f)=\min\{ 1-\sum_{j=1}^L \frac{n_j}{\beta_j n}+\sum_{j=1}^L\frac{n_j}{\beta_j n}R_j^{\ell_{0/1}}(f),\sum_{j=1}^L \frac{n_j}{\alpha_j n} R_j^{\ell_{0/1}}(f) \}$. 
\end{proposition}

From the definition of LAB-CVaR in Eq. \eqref{eq-lab}, we can see that LAB-CVaR degrades to $\alpha$-CVaR when $\alpha_j=\alpha$ and $\frac{1}{\beta_j n}\rightarrow 0$. Then according to Proposition \ref{inverse}, we get the same result with \cite{boostedCVaR} that $\alpha\text{-CVaR}^{\ell_{0/1}}(f)=\min\{ {1,\frac{1}{\alpha}}R^{\ell_{0/1}}(f) \}$, which reveals that the minimizer of ERM is also the minimizer of $\alpha\text{-CVaR}^{\ell_{0/1}}(f)$, and means $\alpha$-CVaR has no better performance than ERM in classification task. 
On the other hand, when the $\alpha_j$ and $\beta_j$ are label-aware, the minimizer of LAB-CVaR$^{\ell_{0/1}}(f)$ is equal to the minimizer of one of the two class-weighted loss,  $\sum_{j=1}^L\frac{n_j}{\beta_j n}R_j^{\ell_{0/1}}(f)$ and $\sum_{j=1}^L \frac{n_j}{\alpha_j n} R_j^{\ell_{0/1}}(f)$. Thus LAB-CVaR has the different minimizer with that of ERM, and could be expected to have better performance than ERM when $\alpha_j$ and $\beta_j$ are properly designed.

Besides the label-aware bound design in LAB-CVaR which helps to overcome the pessimistic result of $\alpha$-CVaR, the another import design in LAB-CVaR is the non-zero lower bound $\frac{1}{\beta_j n}$. Because of the non-zero lower bound, some samples are re-weighted by the upper bound and the others are re-weighted by the lower bound, thus LAB-CVaR takes advantage of all the samples and at the same time discriminatingly treats the samples in the same class. In the following, we theoretically design the optimal upper and lower bounds aiming to minimize the LAB-CVaR loss.



\begin{proposition} \label{bound}
With probability at least $1-\delta$, LAB-CVaR loss is bounded by
\begin{equation}
    \sum_{j=1}^L\frac{1}{\beta_j}\sqrt{n_j}\frac{4L*C(\mathcal{F})}{n} \lesssim \mathbb{E}_{P_{x,y}}\text{LAB-CVaR}^{\ell_{0/1}}(f) \lesssim \sum_{j=1}^L\frac{1}{\alpha_j}\sqrt{n_j}\frac{4L*C(\mathcal{F})}{n},
\end{equation}
where we use $\lesssim$ to hide some constant factors and low order terms, $\mathcal{F}$ is the family of hypothesis class, and $C(\mathcal{F})$ is some proper complexity measure of $\mathcal{F}$.
\end{proposition}

Assume the weighted sum of $\alpha_j$ is constrained by the form that $\sum_{j=1}^Ln_j^{k_1}\alpha_j=\tau_1$, where $\tau_1\in\mathbb{R}^+$, $k_1$ is a variable exponent, then in order to minimize the upper bound of LAB-CVaR in Proposition \ref{bound}, $\alpha_j$ should be proportional to $n_j^{1/4-k_1/2}$, because according to Cauchy–Schwarz inequality, $\sum_{j=1}^L\frac{1}{\alpha_j}\sqrt{n_j}=\frac{1}{\tau_1}(\sum_{j=1}^Ln_j^{k_1}\alpha_j)(\sum_{j=1}^L\frac{1}{\alpha_j}\sqrt{n_j})\geq\frac{1}{\tau_1}(\sum_{j=1}^L n_j^{1/4+{k_1}/2})^2$, if and only if $n_j^{k_1}\alpha_j \propto \frac{1}{\alpha_j}\sqrt{n_j}$. Similarly, assume $\sum_{j=1}^Ln_j^{k_2}\beta_j=\tau_2$, we have $\beta_j\propto n_j^{1/4-k_2/2}$ to minimize the lower bound of LAB-CVaR. Then the optimal $\alpha_j$ and $\beta_j$ are
\begin{equation}\label{optimal-alpha}
    \alpha_j^*=\tau_1\frac{n_j^{1/4-k_1/2}}{\sum_{j=1}^L n_j^{1/4+k_1/2}}, \text{    } \beta_j^*=\tau_2\frac{n_j^{1/4-k_2/2}}{\sum_{j=1}^L n_j^{1/4+k_2/2}},
\end{equation}
where $k_1<\frac{1}{2}$ and $k_2<\frac{1}{2}$ ensure the classes with smaller sample size have larger weight bounds, and $\alpha_j^* < \beta_j^*$ ensures the upper bound is larger than the lower bound. Then the corresponding LAB-CVaR with the optimal bounds is
\begin{equation} \label{optimal}
    \text{LAB-CVaR}^{\ell*}(f) = \max_{\bm{w}\in \Delta_n,\frac{1}{\beta_j^*n} \leq w_i \leq \frac{1}{\alpha_j^*n}} \sum_{i=1}^n w_i\ell(f(x_i),y_i).
\end{equation}

Eq. \eqref{optimal} is a linear programming problem which can be easily solved by the simplex algorithm \cite{simplex}. Further, in our experiment section, we set $k_1 = k_2$. On one hand, this can reduce the hyper-parameter number. On the other hand, it will be easy to ensure $\alpha_j<\beta_j$ just by a simple way, setting $\tau_1=\eta\tau_2$ with $0<\eta<1$. 

\subsection{Label-aware bounded CVaR with logit adjustment}\label{sec:logit}
In practice, the zero-one loss is non-convex and non-continuous, so we usually use a surrogate loss instead of the zero-one loss at training time for the optimization tractability, such as the softmax cross-entropy loss for multiclass classification. A softmax cross-entropy loss over the sample $(x_i,y_i)$ could be written as
\begin{equation}
\begin{aligned}
    \ell_{sc}(f(x_i),y_i) = - \log[\frac{\exp(f_j(x_i))}{\sum_{j\prime\in[L]} \exp(f_{j\prime}(x_i))}] 
    =\log[1+\sum_{j\prime \neq j}\exp(f_{j\prime}(x_i)-f_{j}(x_i))],
\end{aligned}
\end{equation}
where $y_i=j$, $f_j(x_i)$ is the logit of the class $j$, namely \emph{logit} refers to the input of the softmax function.

From Eq. \eqref{optimal}, we can see that LAB-CVaR is a special re-weighting method in which the loss weight is variable during the adversarial training. Without loss of generality, we assume $n_1\leq n_2\leq\cdots\leq n_L$. Then in our LAB-CVaR, the largest weight ratio between two samples is  $\frac{\beta_L^*}{\alpha_1^*}$ which is very large when the labels are extremely imbalanced. Some recent work \cite{cao2019learning,effect-number} find that the large weight ratio can cause difficulties and instability in optimization, which is also observed in our experiments (see Figure \ref{fig:vs}). In this section, we try to reduce the influence of the loss weight in optimization by combining the logit adjustment technique \cite{menon2020long} with our LAB-CVaR.


\begin{proposition} \label{logit}
Define the weighted softmax cross-entropy loss with the logit adjustment as:
\begin{equation}
\begin{aligned}
    R^{\ell_{sc}}_{logit}(f, \pi)
    &= \sum_{i=1}^n 
    \pi_j w_i\log[1+\sum_{j\prime \neq j}\exp(f_{j\prime}(x_i)+\log\pi_{j\prime}-f_{j}(x_i)-\log\pi_{j})].
\end{aligned}
\end{equation}
Then for any $\pi\in\mathbb{R}_+^L$, the loss $R_{logit}^{\ell_{sc}}(f, \pi)$
is consistent with the original weighted softmax loss $\sum_{i=1}^n w_i\ell_{sc}(f(x_i),y_i)$, namely these two losses have the same optimal minimizer.
\end{proposition}

Let $s_1=(x_{i_1},y_{i_1})$ and $s_2=(x_{i_2},y_{i_2})$ be  two samples in the most minority class \emph{1} and the most majority class $L$ respectively. And let the logit $f_j(x)=W_j^T \Phi(x)$, where $W_j$ is the classifier for class $j$, $\Phi(x)$ is the feature extractor. In our opinion, the large weight of $s_1$ makes the sample have more influence on the classifier $W_t$ ($1<t<L$) to push itself away from the class $t$, however, this may make the class $t$ closer with $s_2$.
From Proposition \ref{logit}, we can see that the influence of the loss weight can be transferred to the logit term, then an intuitive solution to reduce the influence of the loss weight is to make the more minority classes have smaller $\pi_j$, which can smooth the $\pi_j w_i$. 

The influence of the loss weight of $(x_i,y_i)$ can be observed by the gradient over the classifier $W_t$. Let $\ell_{log}(f(x_i),y_i)=\pi_j w_i\log[1+\sum_{j\prime \neq j}\exp(f_{j\prime}(x_i)+\log\pi_{j\prime}-f_{j}(x_i)-\log\pi_{j})]$, then the norm of the gradient of $\ell_{log}(f(x_i),y_i)$ over the $W_{t}$ with $\pi=[\pi_1,...,\pi_n]$ is
\begin{equation} \label{gradient}
\begin{aligned}
    g(x_i,y_i,\pi,t)=\|\frac{\partial \ell_{log}(f(x_i),y_i)}{\partial W_t}\|= \pi_{j}w_i\frac{\pi_t\exp({f_{t}(x_i)})}{\sum_{j'\in[L]}\pi_{j'}\exp({f_{j'}(x_i)})}\|\Phi(x_i)\|,\text{    } y_i\neq t.
\end{aligned}
\end{equation}

Through a simple proof, we have that, if the more minority class has the smaller $\pi_j$, this intuitive solution indeed reduces the influence of the loss weight. The simple proof is that, when $\pi_1\leq\cdots\leq\pi_L$, $\frac{g(x_{i_1},y_{i_1},\pi,t)}{g(x_{i_2},y_{i_2},\pi,t)}=\frac{\pi_1 w_{i_1}}{\pi_L w_{i_2}}\frac{\pi_t\exp({f_{t}(x_{i_1})})}{\sum_{j'\in[L]}\pi_{j'}\exp({f_{j'}(x_{i_1})})}\frac{\sum_{j'\in[L]}\pi_{j'}\exp({f_{j'}(x_{i_2})})}{\pi_t\exp({f_{t}(x_{i_2})})}\frac{\|\Phi(x_{i_1})\|}{\|\Phi(x_{i_2})\|} \leq \frac{\pi_1 w_{i_1}}{\pi_L w_{i_2}}\frac{\pi_t\exp({f_{t}(x_{i_1})})}{\pi_1\sum_{j'\in[L]}\exp({f_{j'}(x_{i_1})})}\frac{\pi_L\sum_{j'\in[L]}\exp({f_{j'}(x_{i_2})})}{\pi_t\exp({f_{t}(x_{i_2})})}\frac{\|\Phi(x_{i_1})\|}{\|\Phi(x_{i_2})\|} = \frac{g(x_{i_1},y_{i_1},1,t)}{g(x_{i_2},y_{i_2},1,t)}$, which indicates that the loss of $(x_{i_1},y_{i_1})$ has relatively smaller gradient after being logit adjusted with $\pi$.
 
In this paper, we adopt the policy that $\pi_j=n\alpha_j^*$ which is the reciprocal of the upper weight bound of each class. Then we have our second approach named \emph{label-aware bounded CVaR with logit adjustment} (LAB-CVaR-logit):
\begin{equation}\label{LAB-CVaR-logit}
\begin{aligned}
   \text{LAB-CVaR-logit}^{\ell_{sc}}(f) 
   =\sum_{i=1}^n n\alpha_j^*w_i\log[1+\sum_{j\prime \neq j}\exp(f_{j\prime}(x_i)-f_{j}(x_i)+\log\frac{\alpha_{j\prime}^*}{\alpha_j^*})],
\end{aligned}
\end{equation}
where $w_i$ is calculated in Eq. \eqref{optimal}, namely we firstly get $w_i$ by LAB-CVaR, and secondly use LAB-CVaR-logit to train the model.

\subsection{Comparison to existing work}
\textbf{LHCVaR \cite{xu2020class}.} LHCVaR is also a label-aware bounded method,
our LAB-CVaR is different from it at the following aspects: 1) There is no lower weight bound design in LHCVaR. Furthermore, the upper weight bound designed in LHCVaR is intuitive and lacks theoretical guarantee, whereas we get the optimal $\alpha_j^*$ by the theoretical analysis. 2) In our work, $\alpha_j$ is just the bound of $w_i$ in class $j$, and $w_i$ is still varied across samples even in the same class. While in LHCVaR, the samples in the same class must have the same $w_i$, which will cause the loss of the ability to discriminatingly treat the samples in the same class.

\textbf{Logit adjustment \cite{menon2020long}.} \cite{menon2020long} studies how the logit adjustment technique can make the loss consistent with the balanced loss, whereas we extend the technique to make the loss consistent with the re-weighted loss. The logit adjustment in \cite{menon2020long} is a special case of our LAB-CVaR-logit, because when $k_1=k_2=-\frac{3}{2}$ and $\tau_1=\tau_2$, then $w_i\propto n_j^{-1}$, and LAB-CVaR degrades to the vanilla class-weighted method. Moreover, there is no study in \cite{menon2020long} for the mechanism that the logit adjustment technique can improve the re-weighted methods.

\section{Experiments}\label{experiemnts}
We evaluate the performance of the proposed LAB-CVaR and LAB-CVaR-logit on the imbalanced version of the real-world natural language processing dataset 
(imbalanced version of IMDB Review~\cite{imdb}) and computer vision datasets (the imbalanced version of CIFAR10, CIFAR100~\cite{cifar} and Tiny ImageNet~\cite{imagenet}). 

\textbf{Baselines.} We compare our methods with baselines in the following 4 categories: 1) The classic Empirical Risk Minimization (ERM), which trains network by unweighted cross entropy loss. 2) Three loss function re-weighting methods: the vanilla re-weighting (Vanilla RW), the Class-Balanced Softmax Cross Entropy Loss using Effective Number (CB RW) ~\cite{effect-number}, and the Focal (Focal RW) Loss~\cite{lin2017focal}. The performance gaps among these re-weighting methods are attributed to the difference in their weight designs. Vanilla RW and CB RW use a constant and unified weight for each class during the training. While, Focal RW uses a loss-dependent weight for each sample (see Table 3 in Appendix B). For all of these methods, we do a rescaling of the weighted loss function to make the weight 1 on average in every minibatch. 3) Two logit adjustment methods: Label-Distribution-Aware Margin (LDAM) Loss~\cite{cao2019learning} and Logit-Adjusted (LA) Loss~\cite{menon2020long}. These methods shift the logits of the samples in each class by a constant before feeding the logits to the softmax function and then the cross entropy loss. The difference between these two methods lies in their designs of the shifting constant (see Table 3 in Appendix B). As shown in their paper, LDAM can significantly benefit from the Deferred Re-Weighting (DRW) strategy. So, we add LDAM+DRW as another baseline. 4) The original CVaR loss and one of its variants Label Heterogeneous CVaR (LHCVaR)~\cite{xu2020class} that dynamically optimizes the class weights while training with only an upper bound of the weights.

\textbf{Datasets.} The IMDB Review is a binary classification dataset. The Cifar10, Cifar100 and Tiny ImageNet are multiclass classification datasets. All four datasets originally have class-balanced training and validation sets. To create the imbalanced version of the IMDB Review, we keep the class-balanced validation set unchanged and randomly remove 90\% of positive reviews in the training set. 
To create the imbalanced version of the Cifar10, Cifar100, and Tiny ImageNet, we again keep the class-balanced validation sets unchanged and follow \cite{cao2019learning,menon2020long} to downsample the training sets according to exponential decay. Specifically, the $j$-th class are downsampled to $round(\lambda^{L-j}n_j)$, and $\lambda$ are designed for each dataset to make the imbalance ratio \footnote{The ratio between the sample size of the last class (the most frequent class) and that of the first class (the least frequent class).} 100. To be consistent with the name in \cite{menon2020long}, imbalanced versions of Cifar10, Cifar100, and Tiny ImageNet are named Cifar10-LT, Cifar100-LT, and Tiny ImageNet-LT, respectively. 

\textbf{Experimental Settings.} 
For the imbalanced IMDB Review, we use the two-layer bidirectional LSTM and the Adam optimizer. For Cifar10-LT and Cifar100-LT, we use the ResNet32 and the SGD optimizer. For Tiny ImageNet, we use the ResNet18 and the SGD optimizer. All results reported in this section are averaged over five independent runs and are in the format of $average\pm standard\ deviation$.
More experimental details can be found in Appendix B.

\begin{table}\small\centering
\caption{The error rate (ER\%) of each class and the balanced error rate (BER\%) on imbalanced IMDB Review. Our method LAB-CVaR-logit achieves the best performance.}
\label{tab:imdb_result}
\begin{tabular}{@{}lr@{$\pm$}lr@{$\pm$}lr@{$\pm$}l@{}}
\toprule
               & \multicolumn{2}{c}{Positive Class ER\%} & \multicolumn{2}{c}{Negative Class ER\%} & \multicolumn{2}{c}{BER\%} \\ \midrule
ERM            & 67.03  & 20.49   & \textbf{3.2}     & 1.22   & 35.12      & 9.78  \\ \midrule
Vanilla RW     & 24.46  & 5.13    & 19.96   & 5.30    & 22.21      & 4.87  \\
CB RW~\cite{effect-number}          & 23.49  & 2.52    & 17.56   & 3.45   & 20.52      & 0.84  \\
Focal RW~\cite{lin2017focal}       & 48.49  & 6.71    & 5.41    & 1.35   & 26.95      & 2.68  \\ \midrule
LDAM~\cite{cao2019learning}           & 50.31  & 1.01   & 3.83    & 0.87    & 27.09      & 0.33  \\
LDAM+DRW~\cite{cao2019learning}       & 46.41  & 3.24   & 4.44    & 1.19   & 25.42      & 1.06  \\
LA~\cite{menon2020long}             & 26.20   & 9.68    & 18.06   & 1.23   & 22.13      & 4.94  \\ \midrule
$\alpha$-CVaR  & 67.19  & 25.40    & 3.79    & 2.33   & 35.49      & 11.63 \\
LHCVaR~\cite{xu2020class}         & 33.56  & 7.77    & 23.06   & 4.27   & 28.31      & 2.98  \\ \midrule
LAB-CVaR       & 23.96  & 2.13    & 16.92   & 1.98   & 20.44      & 0.92  \\
LAB-CVaR-logit & \textbf{22.63}  & 0.75    & 17.42   & 0.35   & \textbf{20.02}      & 0.38  \\ \bottomrule
\end{tabular}
\end{table}

\subsection{Results and analysis}

For the imbalanced IMDB Review, the error rate (ER\%) of each class on the validation set and the balanced error rate (BER\%) on the entire validation set are reported  in the Table \ref{tab:imdb_result}. For the Cifar10-LT and Cifar100-LT, the balanced error rate (BER\%) over the entire validation set and the error rate of the worst class (WER\%) in the validation dataset are reported in the Table \ref{tab:main_result}. For the Tiny ImageNet-LT, because the WERs of all algorithms are 100.00\%, besides the balanced error rate (BER\%) over the entire validation set, the error rate of the worst 20 classes (WER20\%) in the validation dataset are reported in the Table \ref{tab:main_result}.



From Table \ref{tab:imdb_result} and Table \ref{tab:main_result}, we can generally see that, under the evaluation of BER and WER, our proposed LAB-CVaR-logit consistently outperforms all the state-of-the-art baselines. On the other hand, our LAB-CVaR does not perform as well as LDAM+DRW and LA in multiclass classification datasets, this is because LAB-CVaR is a re-weighting method which has the optimization problem as aforementioned in Section \ref{sec:logit}, but LAB-CVaR still performs better or comparably to all of the re-weighting baselines. Overall, those results verify the effectiveness of our proposed methods.

On the binary classification dataset IMDB Review, the methods (e.g. ERM) overfitting on the negative class tend to have higher and unstable ER on the positive class. Our methods, though with higher negative class ERs, achieve the best positive class ERs and average BERs with small variances.

\begin{table}\small
\caption{Performance on CIFAR10-LT, CIFAR100-LT and Tiny ImageNet-LT measured by the balanced error rate (BER\%) and the error rate of the worst class (WER\%). Because all the methods have 100.00\% WER on the Tiny ImageNet-LT, the error rate of the worst 20 classes (WER20\%) is further evaluated. Our method LAB-CVaR-logit achieves the best performance.}
\label{tab:main_result}
\begin{tabular}{@{}lr@{$\pm$}lr@{$\pm$}lr@{$\pm$}lr@{$\pm$}lr@{$\pm$}lr@{$\pm$}l@{}}
\toprule
\multicolumn{1}{l}{} & \multicolumn{4}{c}{CIFAR10-LT}                       & \multicolumn{4}{c}{CIFAR100-LT}                      & \multicolumn{4}{c}{Tiny ImageNet-LT}                   \\ \cmidrule(l){2-13} 
\multicolumn{1}{l}{} & \multicolumn{2}{c}{BER\%} & \multicolumn{2}{c}{WER\%} & \multicolumn{2}{c}{BER\%} & \multicolumn{2}{c}{WER\%} & \multicolumn{2}{c}{BER\%} & \multicolumn{2}{c}{WER20\%} \\ \midrule
ERM                  & 30.45       & 0.70      & 51.78       & 2.67      & 64.42       & 0.45      & 100.00      & 0.00      & 77.11       & 0.38      & 99.80        & 0.15       \\ \midrule
Vanilla RW           & 32.59       & 1.36      & 50.42       & 3.88      & 73.12       & 1.62      & 99.00       & 0.00      & 81.17       & 0.34      & 99.10        & 0.08       \\
CB RW~\cite{effect-number}                & 29.07       & 1.26      & 40.86       & 2.33      & 64.35       & 0.80      & 99.00       & 0.00      & 77.01       & 0.27      & 99.90        & 0.05       \\
Focal RW~\cite{lin2017focal}             & 30.28       & 0.43      & 53.02       & 2.43      & 62.21       & 0.69      & 99.80       & 0.40      & 76.80       & 0.41      & 99.90        & 0.16       \\ \midrule
LDAM~\cite{cao2019learning}                 & 29.68       & 0.20      & 49.82       & 3.61      & 64.16       & 0.50      & 100.00      & 0.00      & 77.08       & 0.30      & 99.80        & 0.26       \\
LDAM+DRW~\cite{cao2019learning}             & 23.38       & 0.41      & 33.38       & 1.29      & 60.38       & 0.39      & 97.40       & 1.36      & 76.58       & 0.23      & 98.50        & 0.35       \\
LA~\cite{menon2020long}                   & 22.53       & 0.37      & 33.58       & 1.69      & 57.91       & 1.04      & 95.60       & 1.02      & 75.69       & 0.12      & 97.90        & 0.27       \\ \midrule
$\alpha$-CVaR                 & 30.41       & 0.34      & 53.30       & 2.85      & 65.80       & 0.73      & 100.00       & 0.00      & 76.96       & 0.28      & 100.00        & 0.00       \\
LHCVaR~\cite{xu2020class}               & 28.06       & 0.55      & 45.14       & 1.55      & 63.32       & 0.44      & 99.80       & 0.40      & 77.15       & 0.20      & 100.00       & 0.00       \\ \midrule
LAB-CVaR               & 26.93       & 0.27      & 40.24       & 1.84      & 62.96       & 0.28      & 99.20       & 0.40      & 75.87       & 0.04      & 99.70        & 0.12       \\
LAB-CVaR-logit         & \textbf{21.82}       & 0.22      & \textbf{32.98}       & 0.78      & \textbf{57.29}       & 0.91      & \textbf{95.40}       & 0.49      & \textbf{75.16}       & 0.23      & \textbf{97.52}        & 0.09       \\ \bottomrule
\end{tabular}
\end{table}

\begin{figure}
	\centering
	\begin{subfigure}[b]{0.32\textwidth}\centering
		\includegraphics[height=11em]{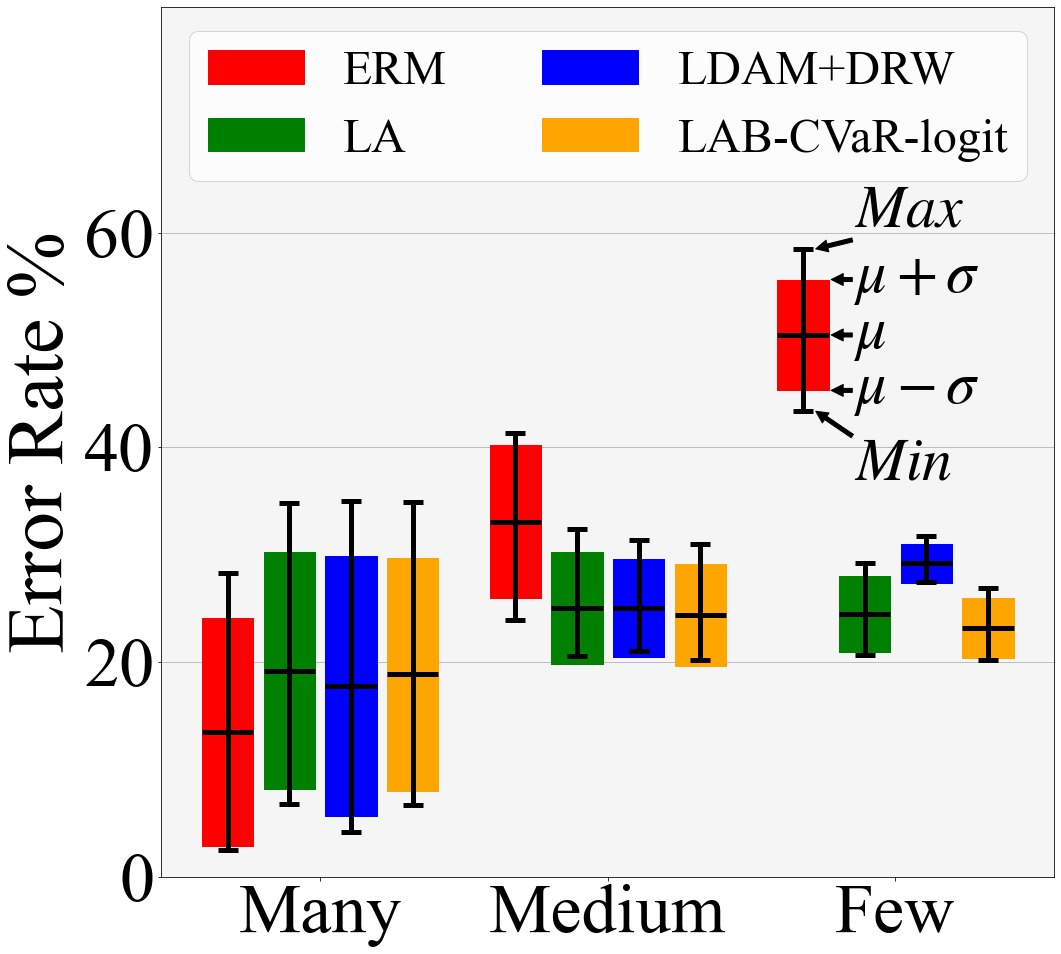}
		\caption{CIFAR10}
		\label{fig:cifar10}
	\end{subfigure}
	\begin{subfigure}[b]{0.33\textwidth}\centering
		\includegraphics[height=11em]{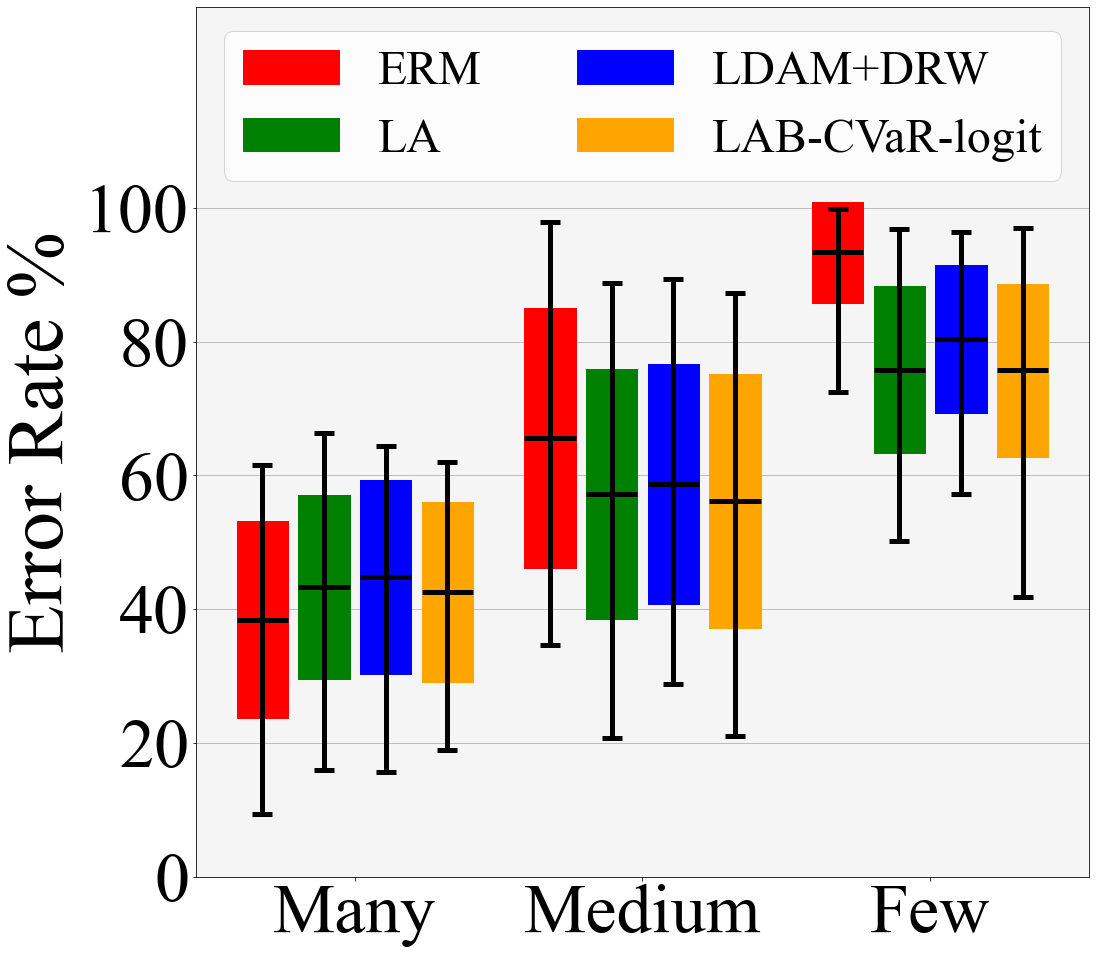}
		\caption{CIFAR100}
		\label{fig:cifar100}
	\end{subfigure}
	\begin{subfigure}[b]{0.33\textwidth}\centering
		\includegraphics[height=11em]{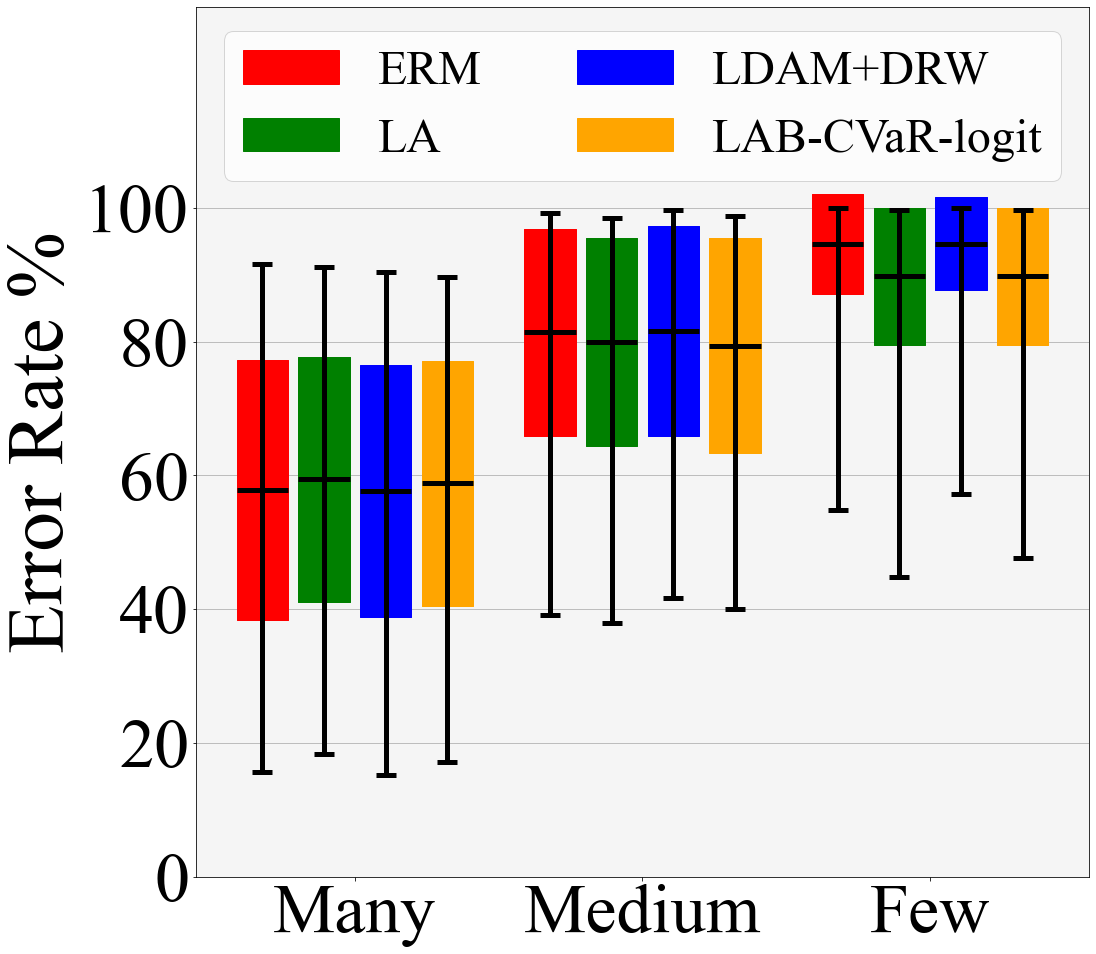}
		\caption{Tiny ImageNet}
		\label{fig:tinyimagenet}
	\end{subfigure}
	\caption{Box plots of the group-wise error rate on imbalanced CIFAR10, CIFAR100 and Tiny ImageNet. Our method LAB-CVaR-logit achieves the best performance when the sample size of the class is Medium and Few. For an explanation of the box plot structure, please refer to the box plot of the \textcolor{red}{ERM} method on the CIFAR10 dataset corresponding to the group Few (in subfigure \ref{fig:cifar10}), where $\mu$ and $\sigma$ refer to the mean and standard deviation, respectively.}
	\label{fig:cw}
\end{figure}
\begin{figure}
	\centering
    \includegraphics[height=12em]{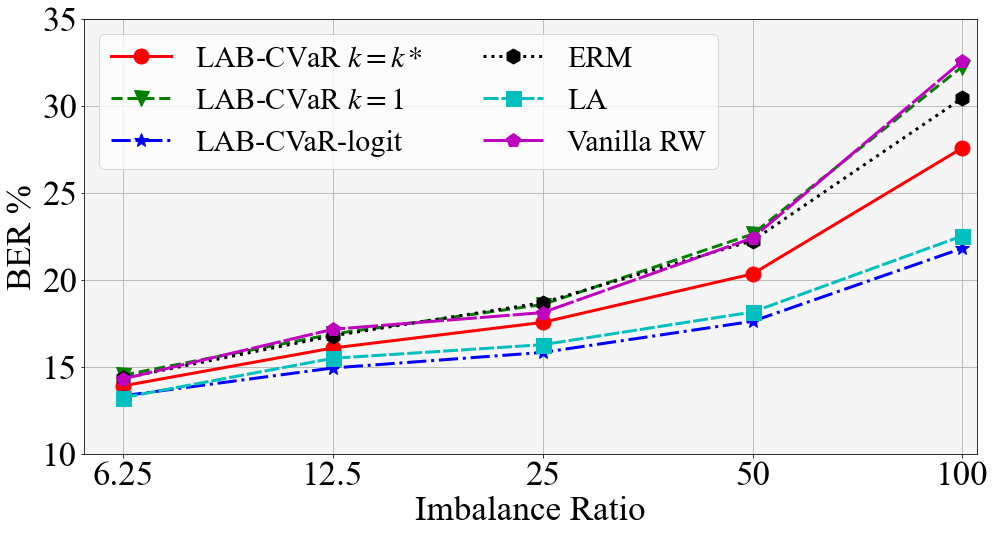}
	\caption{
	The balanced error rate (BER) of some re-weighting methods and logit adjustment methods on imbalanced CIFAR10 for different imbalance ratio. Base on the results, we analyze the optimization problem of re-weighting methods. Here, $k\triangleq1/4-k_1/2=1/4-k_2/2$ is the exponent of the numerator in Eq. \eqref{optimal-alpha}.}
	\label{fig:vs}
\end{figure}
\textbf{Ablation study.}
The results of LAB-CVaR-logit, LAB-CVaR, and $\alpha$-CVaR in the Table \ref{tab:imdb_result} and Table \ref{tab:main_result} constitute an ablation study. The effectiveness of our label-aware bound design is proved by the phenomenon that LAB-CVaR outperforms $\alpha$-CVaR, systematically. The effectiveness of logit adjustment in our method is proved by the phenomenon that LAB-CVaR-logit outperforms LAB-CVaR, systematically. Furthermore, the results of the $\alpha$-CVaR in the Table \ref{tab:imdb_result} and Table \ref{tab:main_result} is not better than that of the ERM, which is consistent with the insight in our Proposition \ref{inverse}. 

\textbf{Group-wise error rate.} 
To investigate where the performance gain of our method compared to baselines comes from. We follow the protocol used in \cite{menon2020long,Kang2020DecouplingRA,Liu2019LargeScaleLR} to group the class into three subsets according to the sample size of the class. Note that the last class has the largest sample size $n_L$ in our experiments. Accordingly, we assign the $j$-th class to the group ``Many'' if $n_j\ge 0.2n_L$, the group ``Medium'' if $0.04n_L\le n_j< 0.2n_L$, the group ``Few'' otherwise. As shown in Figure \ref{fig:cw}, for all three datasets, ERM has the lowest error rate in the ``Many'' group, and our method has the lowest error rate in the ``Medium'' and ``Few'' groups.

\textbf{Performance under different imbalance ratios.}
To analyze how the performances of the methods change under different imbalance levels, we synthesize imbalanced CIFAR10 datasets with different imbalance ratios for further comparison. 
As shown in Figure \ref{fig:vs}, our LAB-CVaR-logit constantly outperforms other methods under all imbalance ratios, and the performance gap between our method and the ERM increases when the imbalance ratio increases. More importantly, the re-weighting methods (Vanilla RW and LAB-CVaR $k=1$) are worse than ERM when the imbalance ratio is high, this phenomenon is consistent with the analysis in Section \ref{sec:logit}, which verifies that the re-weighting method has optimization problems in the highly imbalanced scenarios. Meanwhile, the corresponding logit adjustment methods (LA and LAB-CVaR-logit) can alleviate the optimization problems of the re-weighting methods. We can also find that the LAB-CVaR $k=k^*$ is better than ERM under all the imbalance ratios, this is because $k^*$ is always less than 1, which could reduce the loss weight ratio. However, LAB-CVaR $k=k^*$ is still worse than the logit adjustment methods.


\section{Conclusion}
We propose two approaches to address the long-tailed learning problem, label-aware bounded CVaR (LAB-CVaR), and LAB-CVaR with logit adjustment (LAB-CVaR-logit). These two approaches significantly outperform the state-of-the-art baselines on a variety of imbalanced real-world datasets. Specifically, the LAB-CVaR theoretically designs the optimal weight bounds to minimize its loss bound, and LAB-CVaR-logit adopts the logit adjustments to alleviate the optimization problem caused by re-weighting. The effectiveness of our label-aware bounded CVaR also inspires the study of the sample-aware bounded CVaR to make the best of the differences among the samples in future work.

\clearpage
\bibliographystyle{plain} 
\bibliography{main}



\newpage \label{appendix}

\title{Appendix of "Robust Long-Tailed Learning via \\ Label-Aware Bounded CVaR"}


\maketitle

\appendix 
\section{Appendix: Theoretical Analyses}
\subsection{Proof of Proposition 4.2}
Base on the definition of $\text{LAB-CVaR}^{\ell_{0/1}}$, we have

\begin{equation}
\begin{aligned}
    \text{LAB-CVaR}^{\ell_{0/1}}(f) &=\max_{\bm{w}\in \Delta_n, \frac{1}{\beta_j n} \leq w_i \leq \frac{1}{\alpha_j n}} \sum_{i=1}^n w_i\mathds{1}\{y_i \neq \max_{y'\in [L]} f_{y'}(x_i)\} \\
    &=\sum_{i=1}^n \frac{1}{\beta_j n}\mathds{1}\{y_i \neq \max_{y'\in [L]} f_{y'}(x_i)\} \\
    &\text{    }+\max_{\bm{w}\in \Delta_n^{'}, 0 \leq w_i \leq (\frac{1}{\alpha_j n}-\frac{1}{\beta_j n})} \sum_{i=1}^n w_i\mathds{1}\{y_i \neq \max_{y'\in [L]} f_{y'}(x_i)\}, 
\end{aligned}
\end{equation}
where $\Delta_n^{'}=\{(w_1,\cdots,w_n): \sum_{i=1}^n w_i=1-\sum_{j=1}^L\frac{n_j}{\beta_j n}\}\in \mathbb{R}^n$.

Because $\max_{\bm{w}\in \Delta_n^{'}, 0 \leq w_i \leq (\frac{1}{\alpha_j n}-\frac{1}{\beta_j n})}  w_i\mathds{1}\{y_i \neq \max_{y'\in [L]} f_{y'}(x_i)\}=(\frac{1}{\alpha_j n}-\frac{1}{\beta_j n})\mathds{1}\{y_i \neq \max_{y'\in [L]} f_{y'}(x_i)\}$ whether $\mathds{1}\{y_i \neq \max_{y'\in [L]} f_{y'}(x_i)\}=(\frac{1}{\alpha_j n}-\frac{1}{\beta_j n})\mathds{1}\{y_i \neq \max_{y'\in [L]} f_{y'}(x_i)\}$ is equal to 1 or 0, and note that, $\text{LAB-CVaR}^{\ell_{0/1}}(f)=1-\sum_{j=1}^L\frac{n_j}{\beta_j n}$ when $\sum_{i=1}^n (\frac{1}{\alpha_j n}-\frac{1}{\beta_j n})\mathds{1}\{y_i \neq \max_{y'\in [L]} f_{y'}(x_i)\}>1-\sum_{j=1}^L\frac{n_j}{\beta_j n}$, then we get 
\begin{equation}
    \begin{aligned}
        \text{LAB-CVaR}^{\ell_{0/1}}(f)
        &= \min\{ 1-\sum_{j=1}^L \frac{n_j}{\beta_j n},\sum_{i=1}^n (\frac{1}{\alpha_j n}-\frac{1}{\beta_j n}) \mathds{1}\{y_i \neq \max_{y'\in [L]} f_{y'}(x_i) \} \\
        &\text{    }+\sum_{i=1}^n \frac{1}{\beta_j n}\mathds{1}\{y_i \neq \max_{y'\in [L]} f_{y'}(x_i)\} \\
        &=\min\{ 1-\sum_{j=1}^L \frac{n_j}{\beta_j n}+\sum_{i=1}^n \frac{1}{\beta_j n}\mathds{1}\{y_i \neq \max_{y'\in [L]} f_{y'}(x_i)\}, \\ 
        &\text{    }\sum_{i=1}^n \frac{1}{\alpha_j n} \mathds{1}\{y_i \neq \max_{y'\in [L]} f_{y'}(x_i) \}   \\
        &=\min\{ 1-\sum_{j=1}^L \frac{n_j}{\beta_j n}+\sum_{j=1}^L\frac{n_j}{\beta_j n}R_j^{\ell_{0/1}}(f), \sum_{j=1}^L \frac{n_j}{\alpha_j n} R_j^{\ell_{0/1}}(f) \}
    \end{aligned}
\end{equation}

\subsection{Proof of Proposition 4.3}
In our label-aware bounded CVaR, $\frac{1}{\beta_j n} \leq w_i \leq \frac{1}{\alpha_j n}$, thus
\begin{equation} \label{start}
\sum_{j=1}^L \frac{n_j}{\beta_j n} R_j^{\ell_{0/1}}(f) \leq \text{LAB-CVaR}^{\ell_{0/1}}(f) \leq \sum_{j=1}^L \frac{n_j}{\alpha_j n} R_j^{\ell_{0/1}}(f).
\end{equation}

Inspired by the proof of the Theorem 2 of \cite{cao2019learning}, we prove the loss bound for each class $j$ separately and then union bound over all the classes. From the proof of the Theorem 1 of \cite{xu2020class}, we can see that, with probability at least $1-\delta/L$,
\begin{equation}
    \frac{n_j}{n}\mathbb{E}_{P_{x,y}} R_j^{\ell_{0/1}}\leq \frac{n_j}{n}{R}_j^{\ell_{0/1}}+4L\mathbb{E}_{P_{x,y}}[\frac{n_j}{n}\mathfrak{R}_j(\Pi_1(\mathcal{ F}))]+\sqrt{\frac{\log\frac{L}{\delta}}{2n}},
\end{equation}
where $\mathcal{F}$ is the family of hypothesis class, $\mathfrak{R}_j(\cdot)$ is the empirical Rademacher complexity of class $j$, $\Pi_1(\mathcal{F})=\{ x \mapsto f_j(x): j\in[L],f\in\mathcal{ F}) \}$ 

From the proof of the Corollary 2 of \cite{xu2020class}, we have $\mathbb{E}_{P_{x,y}}[\frac{n_j}{n}\mathfrak{R}_j(\mathcal{F})]\leq C(\mathcal{F})\frac{\sqrt{n_j}}{n}$, where $C(\mathcal{F})$ is some proper complexity measure of $\mathcal{F}$, then
\begin{equation}
    \frac{n_j}{n}\mathbb{E}_{P_{x,y}} R_j^{\ell_{0/1}}\leq \frac{n_j}{n}{R}_j^{\ell_{0/1}}+4L\frac{\sqrt{n_j}}{n}C(\mathcal{F})+\sqrt{\frac{\log\frac{L}{\delta}}{2n}}.
\end{equation}

So  $\mathbb{E}_{P_{x,y}}[\sum_{j=1}^L \frac{n_j}{\alpha_j n} R_j^{\ell_{0/1}}(f)]\leq \sum_{j=1}^L \frac{n_j}{\alpha_j n} R_j^{\ell_{0/1}}(f)+\sum_{j=1}^L \frac{1}{\alpha_j}\sqrt{n_j}\frac{4L*C(\mathcal{F})}{n}+\sqrt{\frac{\log\frac{L}{\delta}}{2n}}\sum_{j=1}^L \frac{1}{\alpha_j}$. Note that, $\frac{n_j}{n}R_j^{\ell_{0/1}}(f)]$ is an arbitrary empirical estimation of $\frac{n_j}{n}\mathbb{E}_{P_{x,y}} R_j^{\ell_{0/1}}$, then
\begin{equation} \label{upper1}
    \mathbb{E}_{P_{x,y}}[\sum_{j=1}^L \frac{n_j}{\alpha_j n} R_j^{\ell_{0/1}}(f)]\leq \sup_{P_{x,y}}\sum_{j=1}^L \frac{n_j}{\alpha_j n} R_j^{\ell_{0/1}}(f)+\sum_{j=1}^L \frac{1}{\alpha_j}\sqrt{n_j}\frac{4L*C(\mathcal{F})}{n}+\sqrt{\frac{\log\frac{L}{\delta}}{2n}}\sum_{j=1}^L \frac{1}{\alpha_j}.
\end{equation}

Remove the low order term $\sqrt{\frac{\log\frac{L}{\delta}}{2n}}\sum_{j=1}^L \frac{1}{\alpha_j}$ and hide the constant term $\sup_{P_{x,y}}\sum_{j=1}^L \frac{n_j}{\alpha_j n} R_j^{\ell_{0/1}}(f)$, we get 
\begin{equation} \label{upper2}
    \mathbb{E}_{P_{x,y}}[\sum_{j=1}^L \frac{n_j}{\alpha_j n} R_j^{\ell_{0/1}}(f)]\lesssim \sum_{j=1}^L \frac{1}{\alpha_j}\sqrt{n_j}\frac{4L*C(\mathcal{F})}{n},
\end{equation}
where we use $\lesssim$ to hide some constant factors and low order terms.

From the Theorem 1.1 of \cite{mendelson2008lower}, we can have $\frac{n_j}{n}\mathbb{E}_{P_{x,y}}R_j^{\ell_{0/1}}\geq \inf_{f\in\mathcal{F}}\frac{n_j}{n}R_j^{\ell_{0/1}}+\frac{c\sqrt{n_j}}{n}$, where $c$ is a constant depending only on $\mathcal{F}$. Because $\frac{4L*C(\mathcal{F})\sqrt{n_j}}{n}<\frac{4L*C(\mathcal{F})}{\sqrt{n}}$, there must exist a constant $\mu\in\mathbb{R}^+$ which satisfies $\frac{c\sqrt{n_j}}{n}\geq \frac{4L*C(\mathcal{F})\sqrt{n_j}}{n}-\mu$ for all the 
class $j$. Then, $\frac{n_j}{n}\mathbb{E}_{P_{x,y}}R_j^{\ell_{0/1}}\geq \inf_{f\in\mathcal{F}}\frac{n_j}{n}R_j^{\ell_{0/1}}+\frac{4L*C(\mathcal{F})\sqrt{n_j}}{n}-\mu$. Similarly with Eq. \eqref{upper1} and Eq. \eqref{upper2}, we get
\begin{equation}
     \sum_{j=1}^L \frac{1}{\beta_j}\sqrt{n_j}\frac{4L*C(\mathcal{F})}{n}\lesssim\mathbb{E}_{P_{x,y}}[\sum_{j=1}^L \frac{n_j}{\beta_j n} R_j^{\ell_{0/1}}(f)].
\end{equation}

Recall Eq. \eqref{start}, we get the result in Proposition 4.3.

By minimizing the bounds $\sum_{j=1}^L \frac{1}{\alpha_j}\sqrt{n_j}\frac{4L*C(\mathcal{F})}{n}$ and $\sum_{j=1}^L \frac{1}{\beta_j}\sqrt{n_j}\frac{4L*C(\mathcal{F})}{n}$, we get the optimal $\alpha^*$ and $\beta^*$, and in fact, the coefficient $4L*C(\mathcal{F})$ does not influence the results of $\alpha^*$ and $\beta^*$.

\subsection{Proof of Proposition 4.4}
The proof is inspired by the proof of Theorem 1 of \cite{menon2020long}.

Denote $\eta_y(x)=P(y|x)$ in the training distribution $P_{x,y}$. Because $P(y|x)\propto \exp(f_y(x))$ could be regarded as estimates of $P_{y|x}$ \citep{menon2020long}, we have $f_y^*(x)=\log \eta_y(x)$.

The class-weighted loss is
\begin{equation}\label{cwERM}
\begin{aligned}
    \mathbb{E}_{P_{x,y}}[\ell_\pi(f(x),y)] = \sum_{y\in[L]} \pi_y P(y)\mathbb{E}_{P_{x|y}}[\ell_{soft}(f(x),y)].
\end{aligned}
\end{equation}

Denote $\tilde{\eta}_y(x)=\tilde{P}(y|x)$ in the class-weighted training distribution $\tilde{P}_{x,y}$. From Eq. \eqref{cwERM}, we can see that $\tilde{P}(y)=\pi_y P(y)$ and $\tilde{P}(x|y)=P(x|y)$. Then we have
\begin{equation}
\begin{aligned}
    \tilde{\eta}_y(x)=\tilde{P}(y|x)=\frac{\tilde{P}(x|y)\tilde{P}(y)}{\tilde{P}(x)}=\frac{{P}(x|y)\pi_y {P}(y)}{\tilde{P}(x)}=\eta_y(x)\pi_y\frac{P(x)}{\tilde{P}(x)}.
\end{aligned}
\end{equation}

So the optimal model of the class-weighted loss is $\tilde{f}_y^*(x)=\log \tilde{\eta}_y(x)=\log \eta_y(x)\pi_y\frac{P(x)}{\tilde{P}(x)}=f_y^*+\log \pi_y +\log \frac{P(x)}{\tilde{P}(x)}$. Because $\log \frac{P(x)}{\tilde{P}(x)}$ does not depend on $y$, we have $\arg\max_{y\in[L]}\tilde{f}_y^*(x)=\arg\max_{y\in[L]} (f_y^*(x)+ \log \pi_y)$. 

If we add a logit $\log\psi_y$ into the class-weighted loss, the corresponding optimal model $\bar{f}_y^*(x)$ satisfies $\bar{f}_y^*(x)+\log\psi_y=\log\tilde{P}(y|x)=\tilde{f}_y^*(x)$. 

Consequently, when $\psi_y=\pi_y$, $\bar{f}_y^*(x)$ is consistent with ${f}_y^*(x)$. 

Note that, $P_{x,y}$ could be any distribution in the above proof, hence we can assume $P_{x,y}$ is the distribution after loss re-weighting by $w_i$, and get the Proposition 4.4.

\section{Appendix: Experiments}
\subsection{The differences in the class weight designs and the shifting constants}
The class weight designs of Vanilla RW, CB RW \footnote{The hyperparameter used in CB RW is denoted by $\beta$ in \cite{effect-number}. To avoid notation conflicts, we denote it by $\gamma$. } and Focal RW, and the shifting constants of the LDAM and LA are shown in the Table \ref{tab:baselines}.
\newcommand{\LdamConstant}{
\left\{
\begin{array}{ll}
    -\frac{C}{n_j^{\frac{1}{4}}}, & j\text{ is the true label}\\
    0, &\text{otherwise}
\end{array}
\right.
}
\begin{table}
\caption{The designs of class weight and shifting constant in our baselines.}\label{tab:baselines}
\centering
\begin{tabular}{@{}ccc@{}}
\toprule
Reweighting Methods      & Weight of the $j$-th Class          & Hyperparameter \\ \midrule
Vanilla RW                & $\frac{1}{n_j}$                     &                 \\[0.4em]
CB RW                   & $\frac{1-\gamma}{1-\gamma^{n_j}}$     & $\gamma\in (0,1)$     \\\cmidrule(l){2-3} 
     & Weight of the $i$-th Sample          & Hyperparameter \\ \cmidrule(l){2-3} 
Focal RW                  & $(1-\ell(f(x_i), y_i))^{\gamma}$        & $\gamma > 0$        \\ \midrule \midrule
Logit Adjustment Methods & Shifting Constant of the j-th Class & Hyperparameter \\ \midrule
LDAM                     & $\LdamConstant$        & $C> 0$             \\[0.4em]
LA          & $-\tau \log n_j$                    & $\tau > 0$          \\ \bottomrule
\end{tabular}
\end{table}

\begin{table}
\centering
	\caption{Hyperparameters and their spaces for random search.}
	\label{tab:hyp}
\begin{tabular}{@{}lcl@{}}
\toprule
Algorithm      & Hyperparameter & Search Space                                         \\ \midrule
CB RW          & $\gamma$        & $\{0.5,0.7,0.8,0.9,0.99,0.999,0.9999\}$              \\ \midrule
Focal RW       & $\gamma$       & $\{0.2,0.5,0.8,1,2,5,10,15\}$                        \\ \midrule
LDAM           & $max_m$        & $\{0.2,0.5,0.8,1,2,5,10,15\}$                        \\ \midrule
LDAM+DRW       & $max_m$        & $\{0.2,0.5,0.8,1,2,5,10,15\}$                        \\
               & DRW Epoch      & int(Total Epoch $\times r$) for $r$ in $\{0.6,0.8\}$ \\ \midrule
$\alpha$-CVaR  & $\alpha$       & $\{0.2,0.5,0.8,1,2,5,10,15\}$                        \\ \midrule
LHCVaR         & $c$            & $\{0.2,0.5,0.8\}$                                    \\
               & $k$            & $\{0.2,0.5,0.8,1,2,5,10,15\}$                        \\ \midrule
LAB-CVaR       & $\tau_1$       & $\{1,2,5\}$                                          \\
               & $\eta$         & $\{1/2,1/3,1/6,1/11,1/16\}$                                    \\
               & $k_1 = k_2$    & $\{0.2,0.5,0.8,1,2,5\}$                              \\ \midrule
LAB-CVaR-logit & $\tau_1$       & $\{1,2,5\}$                                          \\
               & $\eta$         & $\{1/2,1/3,1/6,1/11,1/16\}$                                    \\
               & $k_1 = k_2$    & $\{0.2,0.5,0.8,1,2,5\}$                              \\ \bottomrule
\end{tabular}
\end{table}

\subsection{Implementation Details}
\textbf{IMDB Review.} For IMDB Review, we train the two-layer bidirectional LSTM with batch size of 128 for 10 epochs. We use Adam optimizer with $\beta = 0.9$ and $\beta_1= 0.999$ for training. The initial learning rate is 0.02 and rescaled by 0.1 at the 5-th epoch. 

\textbf{Cifar10 and Cifar100.} For Cifar10 and Cifar100, we train the ResNet32 with a batch size of 128 for 200 epochs. We use stochastic gradient descent with momentum of 0.9, weight decay of $2 \times 10^{-4}$ for training. The initial learning rate is 0.1 and rescaled by 0.01 at the 160-th and 180-th epoch. To be augmented, the training images are horizontally flipped with probability 0.5, padded 4 pixels on each side and randomly cropped to 32 $\times$ 32.

\textbf{Tiny ImageNet.} For Tiny ImageNet, we train the ResNet18 with a batch size of 128 for 120 epochs. We use stochastic gradient descent with momentum of 0.9, weight decay of $2 \times 10^{-4}$ for training. The initial learning rate is 0.1 and rescaled by 0.1 at the 90-th epoch. To be augmented, the training images are horizontally flipped with probability 0.5, padded 8 pixels on each side and randomly cropped to 64 $\times$ 64.

\begin{figure}
	\centering
	\begin{subfigure}[b]{0.24\textwidth}\centering
		\includegraphics[height=6.5em]{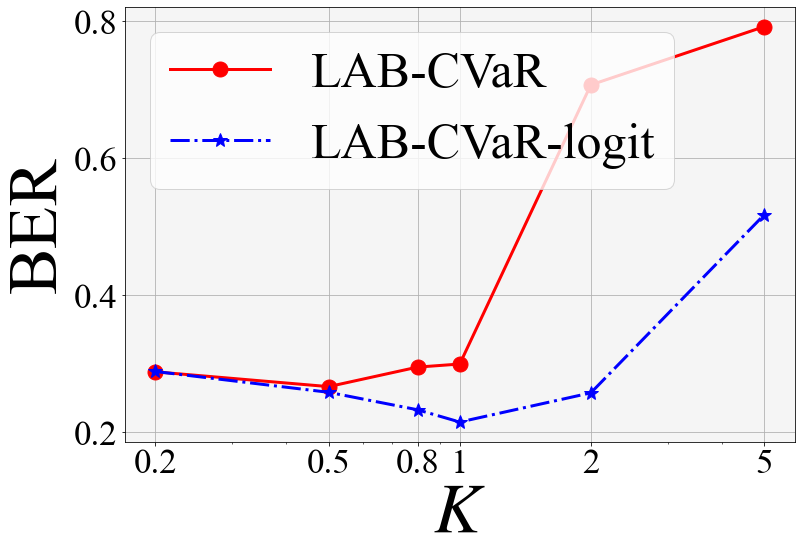}
		\caption{CIFAR10-LT}
		\label{fig:k:cifar10}
	\end{subfigure}
	\begin{subfigure}[b]{0.24\textwidth}\centering
		\includegraphics[height=6.5em]{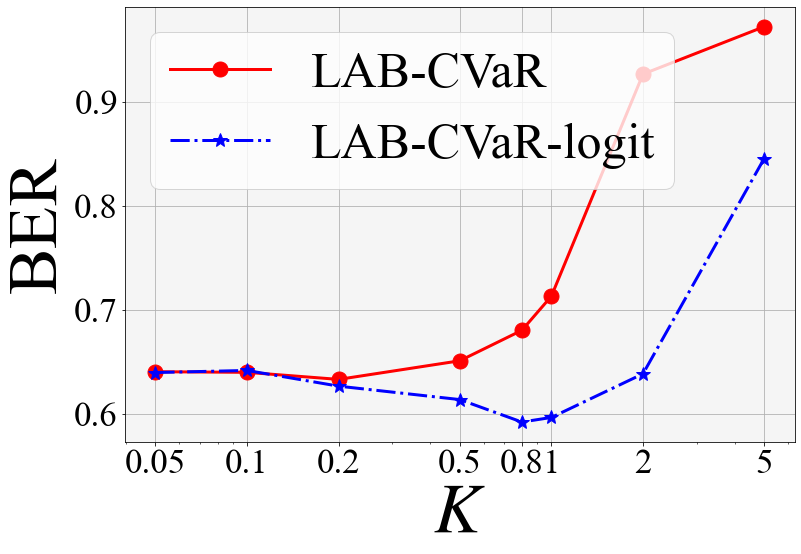}
		\caption{CIFAR100-LT}
		\label{fig:k:cifar100}
	\end{subfigure}
	\begin{subfigure}[b]{0.24\textwidth}\centering
		\includegraphics[height=6.5em]{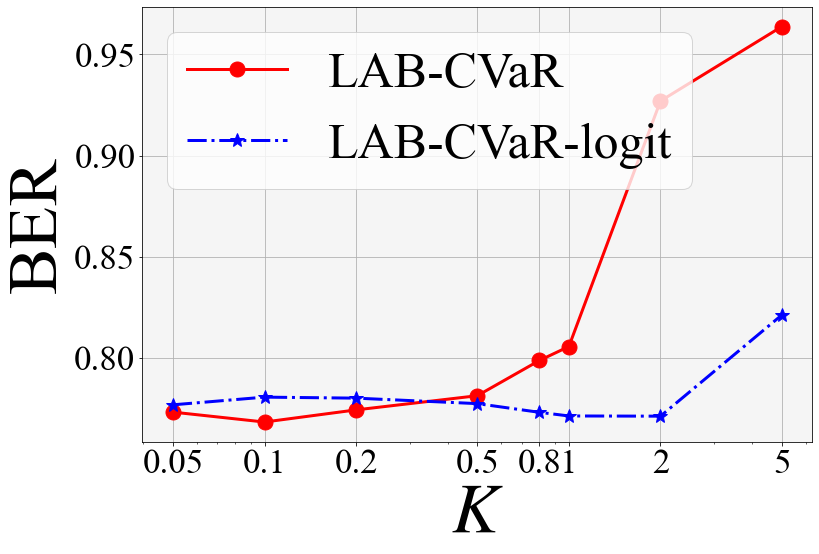}
		\caption{Tiny ImageNet-LT}
		\label{fig:k:tinyimagenet}
	\end{subfigure}
	\begin{subfigure}[b]{0.24\textwidth}\centering
		\includegraphics[height=6.5em]{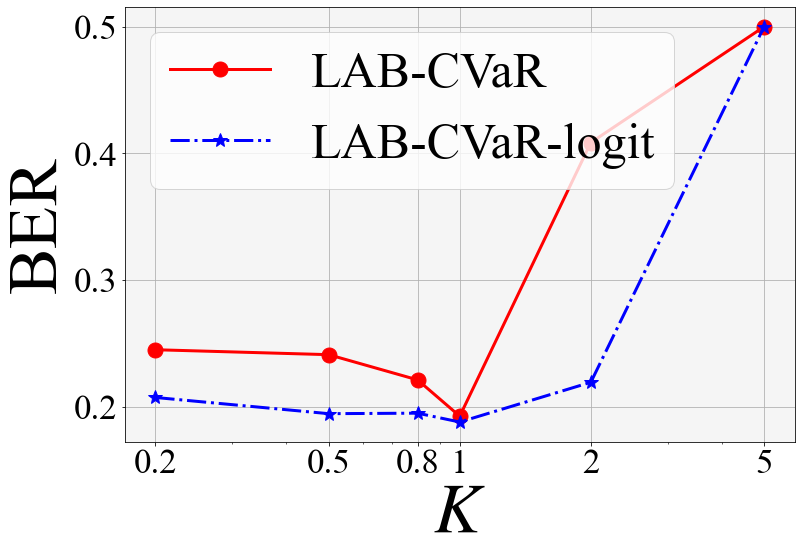}
		\caption{Imbalanced IMDB}
		\label{fig:k:tinyimagenet2}
	\end{subfigure}
	\caption{Performance of LAB-CVaR and LAB-CVaR-logit with different $k$ on the CIFAR10-LT, Cifar100-LT, Tiny ImageNet-LT and imbalanced IMDB Review.}
	\label{fig:k}
\end{figure}

The hyperparameter search space is shown in Table \ref{tab:hyp}. Based on our experimental observation (see Figure \ref{fig:k}), for LAB-CVaR and LAB-CVaR-logit, we would like to suggest choose the exponent $k$ less than 1 and close to 1, respectively. 

\end{document}